\documentclass[sigconf,nonacm]{acmart}
\usepackage{amsmath}
\usepackage{appendix}
\usepackage{graphicx}
\usepackage{booktabs}
\usepackage{array}
\usepackage{float}
\usepackage{multirow}
\usepackage{geometry}
\usepackage{microtype}
\usepackage{ragged2e}
\usepackage[T1]{fontenc}
\usepackage{lmodern}
\usepackage{xurl}
\usepackage{tabularx}
\title{Small Language Models for Inference-Capability Routing: Effects of Model Capacity and Prompt Optimization on Downstream Quality and Efficiency}

\author{Sailendra Kolluru}
\authornote{These authors contributed equally to this work.}
\affiliation{%
  \institution{CORTEXA Innovations Pvt Ltd.}
  \city{Hyderabad}
  \country{India}
}
\email{sailendra.kolluru.int@cortexa.global}

\author{Maheshkumar Duvvarapu}
\authornotemark[1]
\affiliation{%
  \institution{CORTEXA Innovations Pvt Ltd.}
  \city{Hyderabad}
  \country{India}
}
\email{Mahesh.duvvarapu@cortexa.global}

\newenvironment{routerprompt}{%
  \begin{list}{}{%
    \setlength{\leftmargin}{0.6em}%
    \setlength{\rightmargin}{0.6em}%
    \setlength{\topsep}{0.5\baselineskip}%
  }%
  \item\relax
  \small\raggedright
  \setlength{\emergencystretch}{1em}%
}{%
  \end{list}%
}

\begin{document}


\begin{abstract}

Optional reasoning, retrieval, web search, and tool use can improve large language model responses but incur unnecessary computation when activated indiscriminately. This study investigates whether prompt optimization can compensate for limited router capacity and whether classification gains translate into downstream efficiency. The proposed framework uses prompt-driven small language models to make four independent capability decisions, allowing multiple capabilities to be selected per query. Qwen2.5 routers ranging from 1.5B to 14B parameters are compared across five prompting strategies, followed by an evaluation of capability-specific optimized configurations using a fixed GPT-OSS-120B downstream model on 127 questions. Under minimal prompting, increasing router capacity does not consistently improve classification performance. Prompt optimization narrows the capacity gap for reasoning and calculator routing, whereas RAG and web search retain larger benefits from higher capacity. Similar classification F1 scores do not necessarily imply equivalent downstream behavior: false-positive activations increase unnecessary computation, while missed capabilities can reduce answer quality. The 14B router achieves a mean BERTScore F1 of 0.8408, compared with 0.8418 when all capabilities are enabled, while reducing average downstream token consumption by 68.7\% (8,837 to 2,764 tokens) and median downstream latency by 80.5\% (9.79 to 1.91 seconds). These savings exclude router overhead, which is measured separately in a different inference environment and increases with router capacity. The findings demonstrate that the relationship between router capacity, prompt design, and downstream efficiency is capability-dependent. Consequently, router selection must consider error profiles and routing overhead alongside classification performance.
\end{abstract}


\maketitle
\justifying

\section{Introduction}
\label{sec:introduction}

Large language models (LLMs) increasingly operate within
inference pipelines that combine generation with retrieval,
reasoning, and external actions. These mechanisms address
different information and computational requirements.
Retrieval-augmented generation (RAG) supplies external
evidence, web search provides access to online information,
and calculator tools support arithmetic execution.
Additional reasoning effort can help with difficult tasks,
although its benefits depend on the input and downstream model.
Studies on adaptive retrieval and selective reasoning therefore
motivate allocating inference resources according to query
requirements rather than enabling every mechanism
uniformly \cite{adaptive_rag,think_when_needed}.

This allocation creates a capability-routing problem:
which inference capabilities should be enabled for a given
query before downstream answer generation? Activating an
unnecessary capability can introduce additional context,
execution steps, and token consumption. Conversely, omitting
a useful capability can deprive the downstream model of
relevant evidence or computational support \cite{adaptive_rag,think_when_needed}. Routing must
therefore be evaluated in terms of both decision quality
and downstream execution consequences. Existing research
addresses several forms of adaptive inference. Adaptive-RAG
selects among retrieval strategies according to question
complexity \cite{adaptive_rag}. Think When Needed selects
between reasoning and direct inference for ranking tasks
\cite{think_when_needed}, while HRBench compares thinking-mode
switching strategies across models and reasoning domains
\cite{hrbench}. Select-then-Solve examines selection among
complete inference paradigms \cite{select_then_solve}.
These studies establish the importance of adaptive inference,
but their routing targets differ from the independent selection
of multiple capabilities for the same query.

A closely related line of work evaluates small language models
(SLMs) as prompt-driven routers. Guarded Query Routing
demonstrates that prompt optimization can improve endpoint
selection and out-of-distribution rejection without updating
model weights \cite{guarded_query_routing}. This motivates a
further question: when an optimized smaller router approaches
a larger router in classification performance, does it also
produce comparable downstream answer quality and computational
requirements? The answer may vary across capabilities.
Moreover, similar aggregate classification scores can conceal
different false-positive and false-negative profiles, resulting
in different capability activation patterns.

This study investigates these questions using Qwen2.5 routers
with 1.5B, 3B, 7B, and 14B parameters \cite{qwen25} under five prompting
strategies. A single model family is used to reduce
cross-family variation when comparing router capacity.
Each router makes four independent binary decisions:
whether to enable additional reasoning effort, RAG,
web search, and calculator access. This formulation permits
multiple capabilities to be enabled for the same query.
In the downstream implementation, reasoning routing selects
high or low reasoning effort, whereas calculator routing
controls tool availability. The downstream model determines
whether to invoke the available calculator.

The evaluation separates router classification from downstream
execution. First, router capacity and prompting strategies
are compared on capability-specific classification tasks.
Next, the selected prompt configuration for each capability
and router size is fixed, and the resulting decisions are
evaluated using a fixed GPT-OSS-120B downstream model.
Routed configurations are compared against
all-capabilities-on, all-capabilities-off, and
annotation-based oracle conditions. Answer quality is
assessed using BERTScore \cite{bertscore} against reference answers, while
downstream token consumption and latency are measured to
characterize execution requirements. Router overhead is
reported separately because routing and downstream generation
were measured in different inference environments.

The results indicate that increasing router capacity does
not consistently improve classification performance under
minimal prompting. Prompt optimization narrows the capacity
gap for reasoning and calculator routing, whereas RAG and
web search retain larger benefits from higher capacity.
Downstream evaluation further demonstrates that comparable
classification performance does not necessarily imply
comparable execution behavior. The 14B configuration achieves
a mean BERTScore F1 of 0.8408, compared with 0.8418 when
all capabilities are enabled, while reducing average
downstream token consumption by 68.7\%. This finding
concerns downstream computation and does not establish
a corresponding reduction in total system cost after
accounting for router overhead.

This study addresses three research questions, each associated
with a corresponding contribution:

\begin{itemize}
    \item \textbf{RQ1:} How does router capacity affect
    classification performance across reasoning, RAG,
    web-search, and calculator-access decisions?
    \textbf{Contribution:} A within-family comparison of four
    Qwen2.5 model sizes under a common baseline prompt,
    examining how sensitivity to router capacity differs
    across capabilities.

    \item \textbf{RQ2:} How do prompting strategies affect
    classification performance, and to what extent can prompt
    optimization compensate for smaller router capacity
    across capability tasks?
    \textbf{Contribution:} A comparative evaluation of five
    prompting strategies across router sizes and capabilities,
    identifying where prompt optimization narrows the
    performance gap between smaller and larger routers.

    \item \textbf{RQ3:} How do the selected router configurations
    affect downstream answer quality, token consumption,
    and latency?
    \textbf{Contribution:} A downstream evaluation using a
    fixed language model that relates routing decisions and
    error profiles to reference-answer similarity, token
    consumption, and latency, with router overhead reported
    separately.
\end{itemize}

Together, these contributions support evaluating capability
routers through their downstream execution consequences
alongside their classification performance.

\section{Related Work}
\label{sec:related_work}

\subsection{Selective Computation in LLM Systems}

The benefit of additional inference-time computation varies
across queries. Some questions can be answered directly,
whereas others benefit from retrieval, extended reasoning,
or interaction with external tools. This variation motivates
routing policies that allocate computation selectively.
For retrieval, \cite{adaptive_rag} train a small language
model to choose among no retrieval, single-step retrieval,
and iterative retrieval. Their approach connects the routing
target to the outcomes of alternative retrieval strategies,
rather than assuming that every question requires external
evidence.

A similar consideration applies to reasoning. In ranking
tasks, \cite{think_when_needed} demonstrate that reasoning
is not uniformly beneficial and use a lightweight router
head to select between direct inference and reasoning.
Their policy accounts for ranking utility and token
consumption, making the decision sensitive to both task
performance and execution cost. The broader evaluation
by \cite{hrbench} also finds that the effectiveness of
thinking-mode switching depends on the model, task domain,
and switching strategy. These findings suggest that the
value of a routing policy cannot be established independently
of the setting in which its decisions are executed.

Routing decisions can also be made after an initial response.
\cite{inhibitory_deliberation} use a concise answer and
associated evidence to determine whether deeper reasoning
is warranted. This differs from the setting considered in
this study, where capability selection occurs before
downstream answer generation. Consequently, the routers
cannot use the quality or confidence of an initial answer
as a decision signal.

\subsection{Routing Targets and Router Design}

Existing systems route at different levels of an inference
pipeline. \cite{deepsieve} route decomposed questions to
appropriate knowledge sources within an agentic RAG system.
\cite{iterative_agentic_rag} instead select among specialized
agents for corrective retrieval, planning, and workflow
execution. Both approaches adapt processing to the query,
but their routing targets are sources or workflows rather
than individual capability activations.

Selection among complete inference strategies offers another
form of adaptation. \cite{select_then_solve} compare six
inference paradigms and find that no single paradigm performs
best across tasks. A learned embedding-based router exploits
this variation by selecting a paradigm before execution.
Their comparison with self-routing also shows that asking
an LLM to choose a strategy does not necessarily produce
a reliable selection policy.

The proposed routing formulation operates at a different
level. Reasoning, RAG, web search, and calculator access
are predicted separately, allowing multiple capabilities
to be enabled for the same query. These decisions are
evaluated independently at the classification stage, but
their downstream effects need not be independent. For
example, an answer may depend on retrieved evidence as
well as the reasoning applied to it. Consequently, a
capability-level classification error cannot be assumed
to produce a fixed change in answer quality or
computational cost.

\subsection{Prompt Optimization and Downstream Evaluation}

The closest methodological work is that of
\cite{guarded_query_routing}, who evaluate small language
models for domain routing and out-of-distribution rejection.
They show that prompt optimization can improve routing
performance without weight updates and can bring some
smaller models close to larger unoptimized models.
They also identify output-format adherence as a practical
source of failure. This makes prompt design relevant to
both classification quality and the reliability of obtaining
an executable routing decision.

The present study builds on this evidence by examining
whether compensation for smaller router capacity varies
across capability decisions and carries through to downstream
execution. Classification performance alone does not resolve
this question. Two routers with similar F1 scores may differ
in which queries they misclassify and in their balance of
false positives and false negatives. These differences can
change the capabilities made available, the context supplied
to the downstream model, and the amount of computation
performed.

Router capacity and prompting strategies are therefore
evaluated within a single model family, followed by
downstream execution using the selected capability
configurations with a fixed language model. This connects
router-level measurements to reference-answer similarity,
downstream token consumption, and latency. Router overhead
is reported separately. The resulting comparison examines
when a smaller, prompt-optimized router remains competitive
after its decisions are executed, rather than treating
comparable classification scores as sufficient evidence
of equivalent system behavior.

\section{Methodology}
\subsection{Experiment Design}

The methodology consists of two primary stages. The first stage evaluates capability routing as a binary classification problem and encompasses RQ1 and RQ2. For each question, the router independently determines whether reasoning, RAG, web search, and calculator/tool execution are required. Each predicted capability label is compared against an expert-reviewed reference annotation. The second stage evaluates the consequences of these routing decisions during downstream generation and corresponds to RQ3. Rather than assessing only the correctness of routing labels, RQ3 executes the selected capabilities and measures the quality and computational characteristics of the resulting downstream responses. This separation between classification and downstream execution is important because a routing error does not necessarily produce a corresponding change in response quality. For example, a false positive may activate an unnecessary capability and primarily increase computation, whereas a false negative may omit required context and reduce response quality.

The routing problem is decomposed into four independent binary classification tasks. For an input query \(q\), the routing vector is represented as:

\[
\mathcal{R}(q)=
\begin{bmatrix}
r_{\mathrm{reason}} \\
r_{\mathrm{rag}} \\
r_{\mathrm{web}} \\
r_{\mathrm{calc}}
\end{bmatrix},
\]

where each value is binary:

\[
r_i \in \{0,1\}.
\]

A value of 1 indicates that the corresponding capability should be enabled, whereas 0 indicates that it should remain disabled. This formulation permits multiple capabilities to be enabled simultaneously. For example, a query may require both reasoning and web search, or RAG together with calculator execution. Using four independent binary decisions also avoids treating every possible capability combination as a separate class. With four binary capabilities, a unified classifier would otherwise require representing up to \(2^4=16\) different configurations.

\begin{figure*}[t]
    \centering
    \includegraphics[width=\textwidth]{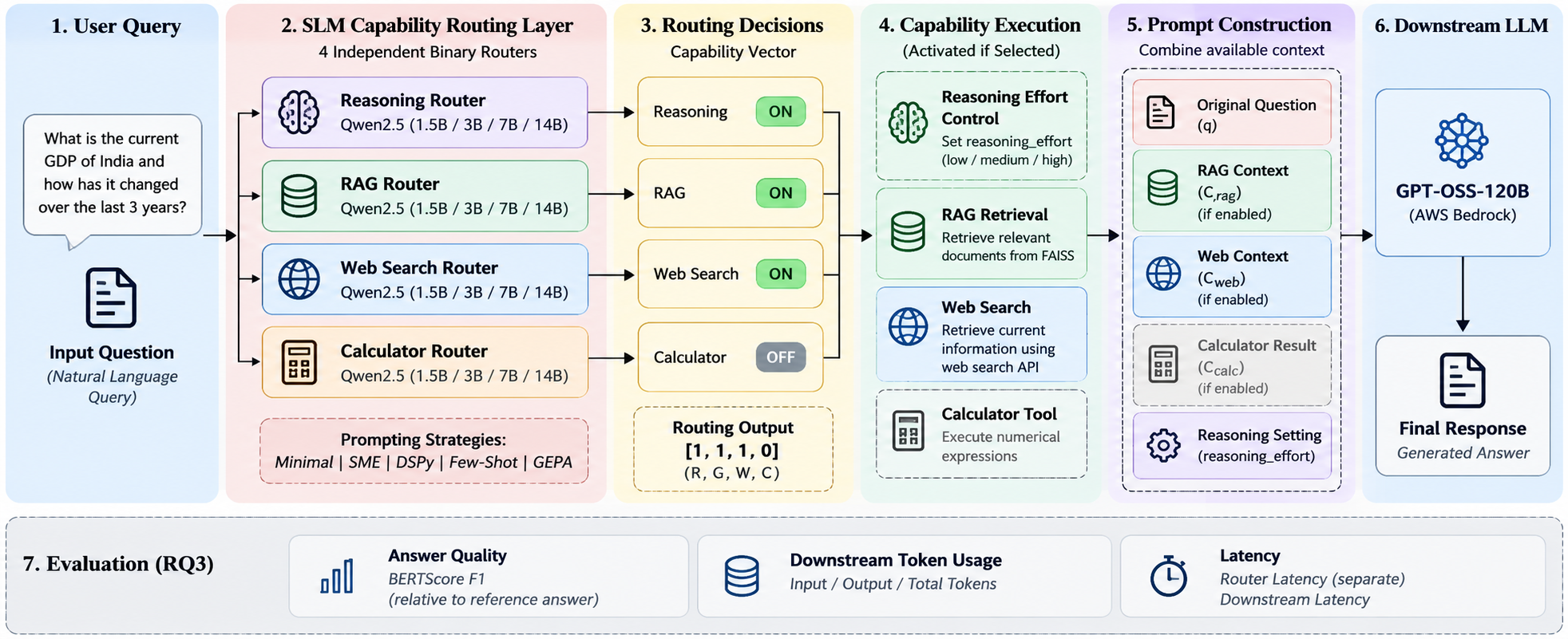}
    \caption{Capability-routing workflow. Four binary predictions
    control downstream reasoning effort, document retrieval,
    web search, and calculator availability. Retrieved context
    and available tools are supplied to a fixed downstream LLM.
    RQ3 evaluates reference-answer similarity, downstream token
    consumption, and latency, with router overhead reported separately.}
    \Description{An input query passes through four capability
    classifiers, producing binary decisions for reasoning,
    document retrieval, web search, and calculator access.
    These decisions configure the inputs and settings of a
    fixed downstream language model. The resulting response
    is evaluated for reference-answer similarity, token
    consumption, and latency, while router overhead is measured
    separately.}
    \label{fig:overall_workflow}
\end{figure*}

============================================================

\subsection{Datasets}

Two evaluation datasets were assembled: a multi-domain capability-classification dataset for RQ1 and RQ2, and a pharmaceutical regulatory question-answering benchmark for RQ3. The datasets were assembled separately but were not fully disjoint. The RQ1/RQ2 dataset combines questions from multiple domains, including pharmaceutical and regulatory content supplied by a subject-matter expert (SME). Source files containing questions and capability annotations were consolidated and normalized into four capability-specific datasets. These records do not constitute a common set of questions annotated for all four capabilities. The RQ3 benchmark was assembled from an initial set of 74 questions provided by the SME, supplemented with additional questions from other datasets supplied by the same SME to broaden capability coverage. Where reference answers were unavailable, candidate answers were generated using GPT-6 and subsequently reviewed and verified by the domain expert. The resulting benchmark contains 127 unique questions, each associated with four capability labels and a reference answer.

Capability annotations for pharmaceutical-domain questions were manually assigned by the SME based on each question's requirements for reasoning, RAG, web search, and calculator/tool use. Domain expertise was particularly important for interpreting pharmaceutical regulatory questions and distinguishing their information and computational requirements.

For a subset of domain-agnostic questions, ChatGPT was used to generate initial capability annotations. These annotations were subsequently reviewed by the SME, who identified and corrected inconsistencies or errors where necessary. The LLM-generated annotations served only as preliminary labels and were not treated as independently established ground truth.

The original RQ1/RQ2 dataset contained 3,573 capability-specific records. A post-experiment consistency audit identified 25 duplicated question-capability pairs with conflicting labels, corresponding to 58 records. All occurrences of these pairs were excluded rather than retaining one of the conflicting labels. The resulting dataset contains 3,515 records: 875 reasoning, 1,003 RAG, 642 web search, and 995 calculator examples. Classification metrics were recomputed using the retained test records.

\begin{table}[H]
\centering
\caption{Final RQ1/RQ2 capability-classification dataset and splits.}
\begin{tabular}{lrrrr}
\toprule
\textbf{Capability} & \textbf{Total} & \textbf{Train} & \textbf{Validation} & \textbf{Test} \\
\midrule
Reasoning & 875 & 94 & 29 & 752 \\
RAG & 1,003 & 100 & 30 & 873 \\
Web Search & 642 & 100 & 30 & 512 \\
Calculator & 995 & 100 & 30 & 865 \\
\midrule
Total & 3,515 & 394 & 119 & 3,002 \\
\bottomrule
\end{tabular}
\end{table}

Training and validation examples were used for prompt development and optimization without updating router model weights. Test examples were used for classification evaluation. The exclusions affected the reasoning training and validation partitions, removing six training examples and one validation example. The training and validation partitions for RAG, Web Search, and Calculator remained unchanged. One excluded reasoning training example had been selected as a DSPy few-shot demonstration for the 3B and 14B routers. The corresponding prompt artifacts therefore reflect development before the consistency audit. Recomputing metrics on the retained test records does not remove the influence of this demonstration. Results for these configurations are interpreted with this limitation. The RQ3 benchmark contains 127 pharmaceutical regulatory questions, each with a reference answer and four capability annotations. It evaluates routing on a domain-specific workload rather than establishing generalization across domains. Some questions also occur in the RQ1/RQ2 datasets. Table~\ref{tab:rq2_rq3_agreement} reports annotation agreement for comparable overlapping question-capability pairs before label harmonization.

\begin{table}[t]
\centering
\caption{Agreement between overlapping RQ2 and RQ3 capability labels before harmonization.}
\label{tab:rq2_rq3_agreement}
\resizebox{\columnwidth}{!}{%
\begin{tabular}{lrrrr}
\toprule
\textbf{Capability} & \textbf{Overlap} & \textbf{Matching} & \textbf{Different} & \textbf{Agreement (\%)} \\
\midrule
Calculator & 83 & 78 & 5 & 93.98 \\
RAG & 87 & 87 & 0 & 100.00 \\
Reasoning & 80 & 53 & 27 & 66.25 \\
Web Search & 87 & 61 & 26 & 70.11 \\
\bottomrule
\end{tabular}%
}
\end{table}

For inconsistent overlapping pairs, RQ3 annotations were harmonized with the retained RQ2 annotations. This changed 58 capability labels across 46 questions: 27 reasoning, 26 web search, and 5 calculator labels. No RAG labels were changed. This procedure establishes consistency between the two annotation sets; it does not, by itself, establish the correctness of the selected reference labels. The final RQ3 ground-truth distribution was:

\begin{table}[H]
\centering
\caption{Final corrected RQ3 ground-truth capability distribution.}
\begin{tabular}{lrr}
\toprule
\textbf{Capability} & \textbf{Required} & \textbf{Not Required} \\
\midrule
Reasoning & 83 & 44 \\
RAG & 97 & 30 \\
Web Search & 10 & 117 \\
Calculator & 29 & 98 \\
\bottomrule
\end{tabular}
\end{table}

Router predictions from the original RQ3 runs were retained and reevaluated against the corrected ground truth. However, because the ORACLE condition directly uses the ground-truth capability vector to control downstream execution, the 46 affected ORACLE runs were rerun using the corrected labels, and their downstream quality metrics were recomputed.

\subsection{Router Classification}

\begin{figure}[t]
    \centering
    \includegraphics[width=\columnwidth]{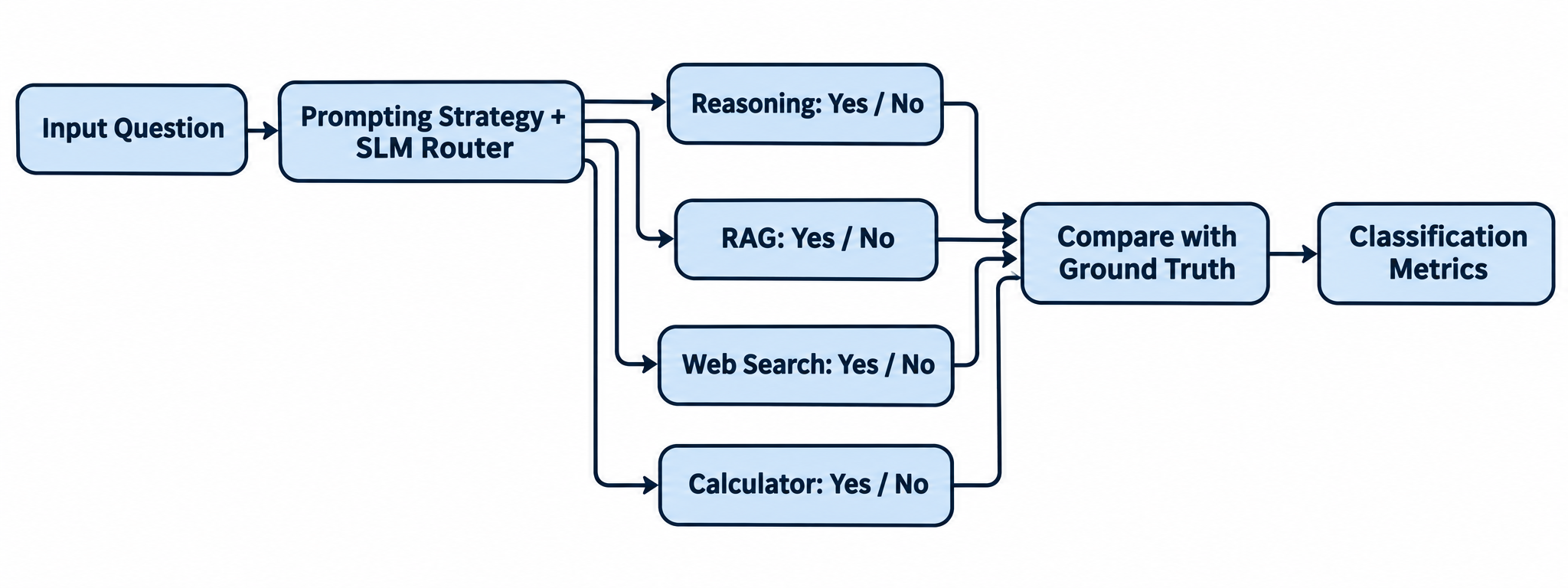}
    \caption{RQ1 and RQ2 capability-classification workflow.
    Each input question is evaluated across router-model
    and prompt-strategy configurations for reasoning, RAG,
    web search, and calculator use, and predictions are
    compared with the corresponding ground-truth labels.}
    \Description{An input question is classified for four
    capabilities: reasoning, retrieval-augmented generation,
    web search, and calculator use. Router model sizes and
    prompting strategies are varied, and the predicted
    capability labels are compared with reference annotations
    to evaluate classification performance.}
    \label{fig:rq2workflow}
\end{figure}

Four Qwen2.5 models were used as routers to evaluate RQ1 and RQ2: Qwen2.5-1.5B, Qwen2.5-3B, Qwen2.5-7B, and Qwen2.5-14B \cite{qwen25}. Using models from the same family reduces architectural variation and enables comparisons of router capacity. All models were executed through Ollama in Google Colab using an NVIDIA T4 GPU. For each model, routing was evaluated independently for reasoning, RAG, web search, and calculator use under five prompting strategies: Minimal Baseline, Descriptive SME, DSPy Baseline, DSPy Few-Shot, and GEPA. The Minimal Baseline contained only the essential classification instruction and output constraint. The Descriptive SME prompt introduced manually designed capability-selection rules, while DSPy-based methods employed structured prompting and optimized few-shot demonstrations \cite{dspy}. GEPA further optimized the routing instructions through a generator-evaluator process \cite{gepa}. RQ1 was evaluated by comparing the four router sizes under the same Minimal Baseline prompt, thereby isolating the effect of model capacity. RQ2 was evaluated by comparing prompting strategies within each router size and capability. Where prompt optimization required training or validation examples, only the designated training and validation partitions were used, while final performance was measured on the corrected held-out test sets. The RQ1/RQ2 workflow is illustrated in Figure~\ref{fig:rq2workflow}.

\subsubsection{Router-Level Evaluation Metrics}

Router performance was evaluated separately for reasoning, RAG, web search, and calculator use. The evaluation considered classification performance, output reliability, and computational overhead.

\paragraph{Classification performance.}
F1 and balanced accuracy were used as the primary classification metrics. F1 summarizes precision and recall for the capability-required class, while balanced accuracy assigns equal weight to sensitivity and specificity:
\[
\mathrm{F1} =
\frac{2TP}{2TP+FP+FN},
\]
\[
\mathrm{Balanced\ Accuracy} =
\frac{1}{2}
\left(
\frac{TP}{TP+FN}
+
\frac{TN}{TN+FP}
\right).
\]
Here, \(TP\), \(TN\), \(FP\), and \(FN\) denote true positives, true negatives, false positives, and false negatives, respectively. Both metrics were reported because class imbalance can obscure routing behavior. In particular, F1 does not account for true negatives, whereas balanced accuracy reflects discrimination across both classes. Accuracy, precision, recall, specificity, and false-positive and false-negative rates were retained as supplementary measures. Confusion counts were also reported to distinguish unnecessary capability activations from missed activations.

\paragraph{Output reliability.}
Unknown predictions and parsing or execution errors were recorded separately from valid binary predictions. Their rates were calculated relative to the total number of evaluated records:
\[
\mathrm{Unknown\ Rate} = \frac{N_{\mathrm{unknown}}}{N},
\qquad
\mathrm{Error\ Rate} = \frac{N_{\mathrm{error}}}{N}.
\]
The number of successfully classified records was retained to make the coverage of the classification metrics explicit.

\paragraph{Computational overhead.}
Router latency was summarized using the mean, median, and 95th percentile. Router token consumption and model runtime statistics were retained where available. These measurements characterize the overhead introduced by capability selection and were reported separately from downstream generation.


\subsection{End-to-End Capability Routing}

RQ3 evaluates how capability-based routing affects downstream answer quality and computational efficiency. Rather than carrying forward a single best router configuration, RQ3 compares four router sizes: 1.5B, 3B, 7B, and 14B. For each router size, the prompting strategy for each capability was selected based on the highest F1 score observed in the RQ2 experiments. Consequently, each router size represents a capability-specific combination of prompting strategies rather than a single prompting method applied uniformly across all four routing decisions. The prompting configurations used in RQ3 are summarized in Table~\ref{tab:rq3_router_prompts}.

\begin{table}[ht]
\centering
\caption{Prompting strategies used for each router size and capability in RQ3.}
\label{tab:rq3_router_prompts}
\small
\setlength{\tabcolsep}{3pt}
\begin{tabular}{lcccc}
\hline
\textbf{Size} & \textbf{Reasoning} & \textbf{RAG} & \textbf{Web} & \textbf{Calculator} \\
\hline
1.5B & Few-Shot & GEPA & GEPA & SME \\
3B   & Few-Shot & GEPA & GEPA & Minimal \\
7B   & Few-Shot & GEPA & GEPA & SME \\
14B  & Few-Shot & GEPA & GEPA & Few-Shot \\
\hline
\end{tabular}
\end{table}

These four router configurations were evaluated as separate routing conditions in the downstream experiment. Because the prompting strategy for each capability was selected using the final RQ2 evaluation results rather than an independent validation set, the RQ3 comparison is interpreted as exploratory rather than as an independently validated estimate of generalization performance. The downstream model was kept fixed across all conditions so that differences in answer quality, token consumption, and latency could be attributed primarily to differences in capability selection. The downstream model was \texttt{openai.gpt-oss-120b-1:0}, accessed through AWS Bedrock. For each input question, the router independently determined whether each of four optional inference capabilities should be enabled: extended reasoning, RAG retrieval, web search, and calculator access. Thus, RQ3 evaluates the downstream behavior of the four optimized router-size configurations alongside the non-routed and oracle baselines.


\subsubsection{Experimental Conditions}

Seven downstream conditions were evaluated: ALL\_OFF, ALL\_ON, ORACLE, ROUTER\_1.5B, ROUTER\_3B, ROUTER\_7B, and ROUTER\_14B. The \textbf{ALL\_OFF} condition disables all optional capabilities and provides a low-computation baseline. The \textbf{ALL\_ON} condition enables all available capabilities for every question and represents a high-computation baseline.

The \textbf{ORACLE} condition uses the benchmark ground-truth capability labels directly as the routing vector. It represents perfect agreement with the benchmark annotations but should not be interpreted as a proven minimum-cost routing policy or an upper bound on downstream answer quality. The four \textbf{ROUTER} conditions use the capability decisions generated by the corresponding Qwen2.5 routers.

For downstream execution, invalid router outputs were handled using a conservative fallback policy. If a router returned an unknown prediction or encountered an execution error, the corresponding capability was enabled for that question. No retry was performed. This ensured that invalid routing outputs did not suppress potentially required capabilities, although such cases could increase downstream computational cost.

\subsubsection{Reasoning Execution}

The reasoning decision controls the reasoning effort assigned to the downstream model. When reasoning is enabled, \texttt{reasoning\_effort} is set to \texttt{high}; otherwise, it is set to \texttt{low}. This setting is passed directly as a downstream inference parameter rather than being appended to the textual prompt.

\subsubsection{RAG Execution}

When RAG is enabled, the input question is used to retrieve relevant information from the experiment's retrieval index. The RAG system uses a FAISS-based vector index \cite{faiss} containing the available reference documents, with sentence embeddings generated using the Sentence-Transformers framework \cite{sentencebert}.

The retrieved passages are converted into contextual information and appended to the downstream prompt. When RAG is disabled, no retrieval operation is performed, and no RAG context is included.

\subsubsection{Web Search Execution}

When web search is enabled, the question is submitted to the Tavily search API. The returned search results are processed into web context and supplied to the downstream model. When web search is disabled, no Tavily request is executed, and no web-search context is added to the prompt.

\subsubsection{Calculator Execution}

When calculator access is enabled, the calculator/tool component is made available to the downstream model, which determines whether a calculation is necessary. If invoked, the calculator executes the relevant mathematical operation and returns the computed result to the downstream model. When calculator access is disabled, the calculator tool is unavailable.

\subsubsection{Downstream Prompt Construction}

\begin{figure}[H]
    \centering
    \includegraphics[width=\linewidth]{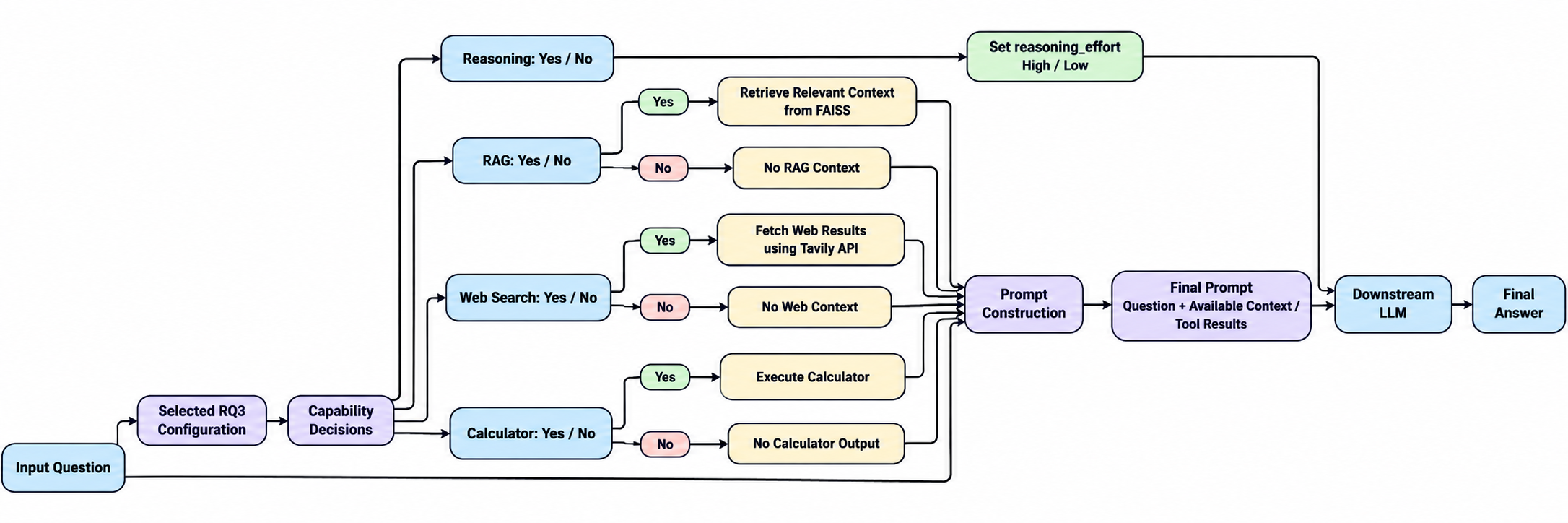}
    \caption{RQ3 end-to-end routing workflow. The selected configuration determines whether reasoning, RAG, web search, and calculator capabilities are enabled before the resulting context is passed to the fixed downstream LLM.}
    \Description{An input question is processed under a
    baseline, oracle, or SLM-routed capability configuration.
    The configuration controls reasoning effort, document
    retrieval, web search, and calculator access for a fixed
    downstream language model. Generated responses are
    evaluated for reference-answer similarity, token
    consumption, and latency.}
    \label{fig:rq3workflow}
\end{figure}

The downstream prompt combines the original question with any context produced by enabled RAG and web-search capabilities. Calculator results are incorporated when the available tool is invoked. Reasoning is handled separately through the model's \texttt{reasoning\_effort} parameter rather than as additional prompt text. The complete RQ3 execution workflow is shown in Figure~\ref{fig:rq3workflow}.

\subsection{RQ3 Evaluation Metrics}

RQ3 evaluates both response quality and computational efficiency. Generated answers were compared with the available reference answers using BERTScore \cite{bertscore}. The primary response-quality metric was BERTScore F1. BERTScore estimates semantic similarity between generated and reference responses using contextual token representations and is less dependent on exact lexical overlap than traditional string-matching metrics. Token consumption was recorded separately for routing and downstream generation. The downstream analysis included input tokens, output tokens, and total downstream tokens. Router token consumption was retained separately to allow examination of routing overhead without combining it with downstream model usage. Latency measurements included router latency, downstream latency, RAG latency, web-search latency, calculator latency, and end-to-end execution characteristics. Mean, median, and 95th-percentile (P95) latency statistics were calculated where appropriate.

\section{Reproducibility and Data Quality}

All routing experiments were executed in Google Colab using an NVIDIA T4 GPU. Because the experimental matrix required a large number of router and downstream inference calls, intermediate predictions and outputs were continuously saved to Excel checkpoints. This allowed interrupted runs to resume from previously completed examples and enabled evaluation to be repeated using stored outputs without rerunning the full inference pipeline. A post-experiment consistency audit identified conflicting labels within the RQ1/RQ2 dataset and inconsistencies between overlapping RQ2 and RQ3 annotations. Conflicting question-capability pairs were excluded, and all affected RQ1/RQ2 classification metrics were recalculated using the corrected test sets. These corrections produced only minor changes in classification metrics and did not alter the main RQ1 capacity trends or the overall prompt-selection patterns observed in RQ2. The audit also identified one excluded reasoning example that had previously been selected as a DSPy Few-Shot demonstration for the 3B and 14B reasoning routers. Because the corresponding prompts and evaluations had already been completed, these configurations were neither regenerated nor rerun. Their results were therefore retained but should be interpreted as exploratory, with this prompt exposure acknowledged as a limitation of the study. For RQ3, overlapping capability labels were harmonized with the corrected RQ2 annotations. Existing router predictions were retained and reevaluated against the corrected ground truth to isolate the effect of label corrections without introducing new router outputs. The ORACLE condition required separate treatment because its downstream capability configuration was determined directly by the ground-truth routing vector. Consequently, the 46 affected ORACLE runs were rerun, after which BERTScore values and downstream summary statistics were recomputed. All subsequent RQ3 analyses were generated using these corrected outputs.

The detailed software, model, and inference configurations are provided in Tables~\ref{tab:experimental_configuration} and~\ref{tab:experimental_configuration_continued} in the Appendix.

\section{Results and Analysis}

All classification metrics were recalculated following the label and example audit. Precision, recall, F1 score, balanced accuracy, and standard accuracy were computed using the conventional confusion-matrix definitions over valid classified predictions; outputs marked as unknown or error were excluded from these calculations. Strict accuracy additionally accounted for invalid outputs by treating unknown and error cases as incorrect predictions, thereby measuring performance across the full set of examples. An output was marked as unknown when the router produced a response that could not be mapped to a permitted classification label, such as ``none of the above.'' Errors were recorded when inference failed to produce a valid prediction, for example because of a token-limit exceedance, timeout, or termination without a usable response. Consequently, the number of examples included in the standard classification metrics may be smaller than the full dataset for configurations with unknown or error outputs, whereas strict accuracy includes these cases in its denominator.

\subsection{RQ1: Effect of Router Capacity}

RQ1 examined how router classification performance varied with model capacity under the Minimal Baseline prompt. Table~\ref{tab:rq1_f1_balacc} reports F1 score and balanced accuracy.

\begin{table}[t]
\centering
\caption{RQ1 performance under the Minimal Baseline prompt.}
\label{tab:rq1_f1_balacc}
\resizebox{\columnwidth}{!}{%
\begin{tabular}{llrrrr}
\toprule
\textbf{Capability} & \textbf{Metric} & \textbf{1.5B} & \textbf{3B} & \textbf{7B} & \textbf{14B} \\
\midrule
\multirow{2}{*}{Calculator}
& F1                & 0.644 & \textbf{0.928} & 0.917 & 0.888 \\
& Balanced Accuracy & 0.644 & \textbf{0.936} & 0.924 & 0.903 \\
\midrule
\multirow{2}{*}{RAG}
& F1                & \textbf{0.678} & 0.602 & 0.196 & 0.579 \\
& Balanced Accuracy & 0.509 & 0.604 & 0.484 & \textbf{0.624} \\
\midrule
\multirow{2}{*}{Reasoning}
& F1                & 0.745 & 0.771 & \textbf{0.809} & 0.768 \\
& Balanced Accuracy & 0.500 & 0.569 & \textbf{0.663} & 0.559 \\
\midrule
\multirow{2}{*}{Web Search}
& F1                & 0.351 & 0.487 & 0.484 & \textbf{0.533} \\
& Balanced Accuracy & 0.500 & 0.712 & 0.707 & \textbf{0.753} \\
\bottomrule
\end{tabular}%
}
\end{table}

Under minimal prompting, increasing router capacity did not produce a consistent improvement in classification performance. The 3B and 7B models matched or exceeded the 14B model on several measures, particularly for calculator and reasoning routing, whereas RAG and web search results were more variable. Overall, model size alone was not a reliable predictor of routing performance.

\begin{table}[H]
\centering
\caption{Unknown predictions under the Minimal Baseline prompt.}
\label{tab:rq1_unknowns}
\begin{tabular}{lrrrr}
\toprule
\textbf{Capability} & \textbf{1.5B} & \textbf{3B} & \textbf{7B} & \textbf{14B} \\
\midrule
Reasoning   & 0 & 0 & 0 & 0 \\
RAG         & 0 & 0 & 0 & 0 \\
Web Search  & 0 & 0 & 0 & 0 \\
Calculator  & 0 & 0 & 0 & 24 \\
\bottomrule
\end{tabular}
\end{table}

An additional difference emerged in output reliability under minimal prompting. While the smaller routers returned valid labels across all four tasks, the 14B model produced 24 unknown outputs for calculator routing, often returning responses such as ``none of the above'' rather than one of the required labels. This suggests that additional model capacity does not necessarily ensure adherence to a strict binary output format under minimal prompting.

\subsection{RQ2: Effect of Prompt Engineering}

Prompting strategy substantially affected routing performance, although the most effective strategy varied by capability. GEPA consistently achieved the strongest results for RAG and Web Search, while DSPy Few-Shot performed best for Reasoning. Calculator routing showed a different pattern, with Minimal, Descriptive SME, and Few-Shot prompts each performing best at different model sizes.

\begin{table}[H]
\centering
\caption{Best RQ2 F1 score for each capability and router size.}
\label{tab:rq2_best_f1}
\begin{tabular}{lrrrr}
\toprule
\textbf{Capability} & \textbf{1.5B} & \textbf{3B} & \textbf{7B} & \textbf{14B} \\
\midrule
Calculator & 0.905 & 0.928 & 0.934 & 0.941 \\
RAG        & 0.712 & 0.657 & 0.870 & 0.955 \\
Reasoning  & 0.871 & 0.865 & 0.855 & 0.913 \\
Web Search & 0.651 & 0.639 & 0.834 & 0.868 \\
\bottomrule
\end{tabular}
\end{table}

Unlike the Minimal Baseline results, for which increasing capacity did not consistently improve performance, the best-prompt configurations showed a clearer capacity effect: the 14B router achieved the highest F1 for all four capabilities. However, the size of this advantage was task-dependent. For reasoning and calculator routing, smaller routers remained close to the 14B model, suggesting that prompt optimization can substantially narrow the capacity gap. In contrast, RAG and Web Search showed larger gains at higher capacities, indicating that prompting could not fully compensate for limited model capacity on these tasks. Overall, prompt optimization substantially improved several routing tasks, particularly RAG and Web Search, but its effect remained capability-dependent. More complex prompting was therefore not universally superior; the strongest configuration depended on both the routing task and model capacity.

\begin{table}[H]
\centering
\caption{Unknown predictions by prompting strategy and router model.}
\label{tab:rq2_unknowns}
\begin{tabular}{lrrrr}
\toprule
\textbf{Method} & \textbf{1.5B} & \textbf{3B} & \textbf{7B} & \textbf{14B} \\
\midrule
Minimal Baseline & 0   & 0 & 0 & 24 \\
Descriptive SME  & 0   & 0 & 0 & 6  \\
DSPy Baseline    & 358 & 6 & 0 & 0  \\
DSPy Few-Shot    & 53  & 6 & 0 & 1  \\
GEPA             & 32  & 6 & 0 & 1  \\
\bottomrule
\end{tabular}
\end{table}

Although optimized prompting improved F1 for several configurations, its effect on output reliability depended strongly on model capacity. Unknown predictions were concentrated in the 1.5B router under DSPy and GEPA, whereas the 7B model produced no unknown outputs across any prompting strategy. The 14B model showed the opposite pattern, with most unknowns occurring under the Minimal Baseline. These findings suggest that prompt complexity and model capacity interact in both classification performance and the reliability of producing valid routing labels.

\begin{figure}[H]
    \centering
    \includegraphics[width=\linewidth]{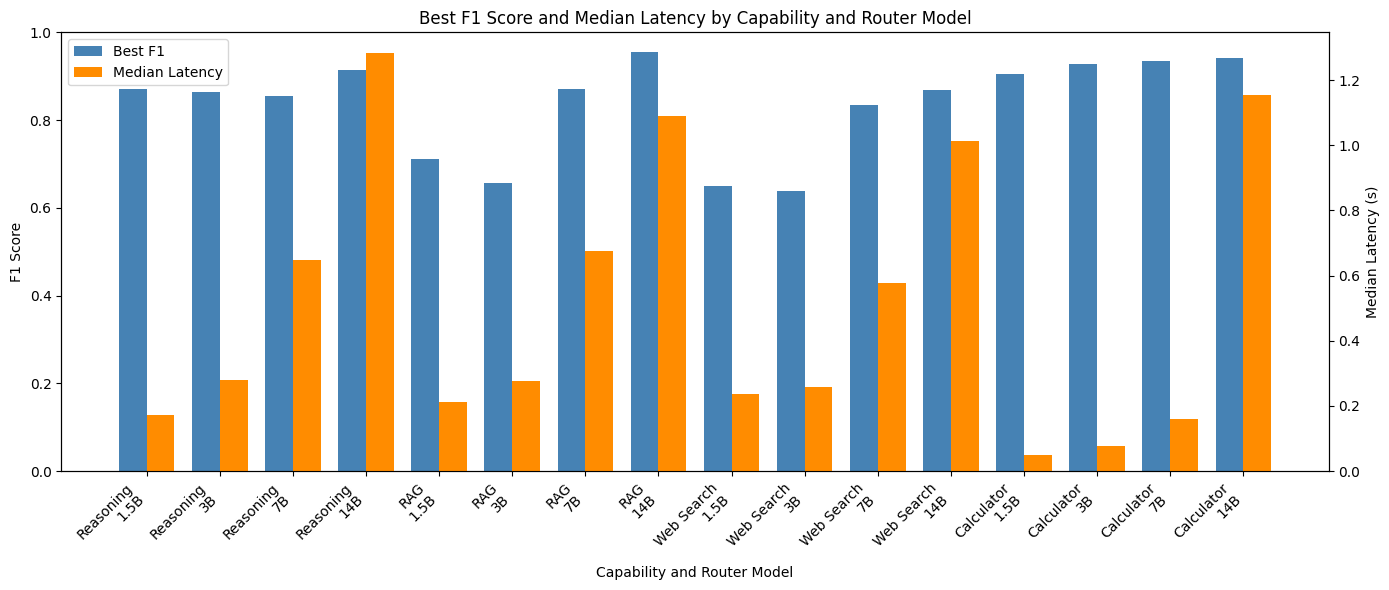}
    \caption{Best F1 score and median router latency for the highest-performing prompt configuration at each capability and model size.}
    \Description{}
    \label{fig:rq2_f1_latency}
\end{figure}

As shown in Figure~\ref{fig:rq2_f1_latency}, the best-performing configurations also exhibited a capability-dependent performance-latency trade-off. For reasoning and calculator routing, smaller routers approached the F1 performance of the 14B model at substantially lower median latency. In contrast, RAG and Web Search showed larger gains at higher capacities, making the additional router latency more justifiable for these tasks.
\subsection{RQ3: Router Performance on the Downstream Benchmark}
Before downstream answer quality and efficiency were evaluated, the router configurations were assessed on the corrected RQ3 benchmark. Table~\ref{tab:rq3_router_performance} reports F1 score and balanced accuracy (BA) for each capability and router size.

\begin{table}[H]
\centering
\caption{Router classification performance on the corrected RQ3 benchmark.}
\label{tab:rq3_router_performance}
\begin{tabular}{llrrrr}
\toprule
\textbf{Capability} & \textbf{Metric} & \textbf{1.5B} & \textbf{3B} & \textbf{7B} & \textbf{14B} \\
\midrule
\multirow{2}{*}{Reasoning}
& F1                & 0.607 & 0.676 & \textbf{0.740} & 0.716 \\
& BA & 0.556 & 0.698 & 0.684 & \textbf{0.755} \\
\midrule
\multirow{2}{*}{RAG}
& F1                & 0.580 & 0.328 & 0.715 & \textbf{0.953} \\
& BA & 0.661 & 0.598 & 0.778 & \textbf{0.908} \\
\midrule
\multirow{2}{*}{Web Search}
& F1                & \textbf{0.243} & 0.143 & 0.118 & 0.194 \\
& BA & \textbf{0.715} & 0.537 & 0.524 & 0.573 \\
\midrule
\multirow{2}{*}{Calculator}
& F1                & 0.446 & 0.667 & 0.792 & \textbf{0.846} \\
& BA & 0.627 & 0.760 & 0.828 & \textbf{0.874} \\
\bottomrule
\end{tabular}
\end{table}

Because reasoning-required questions are relatively prevalent in the corrected RQ3 benchmark (83 of 127 questions), an analytical always-positive classifier would obtain an F1 score of approximately 0.790, which exceeds the reported Reasoning F1 values of all four routers. This baseline does not account for the computational cost of unnecessarily enabling high reasoning effort, but it provides important context for interpreting the router F1 scores.

Router performance varied considerably across capabilities. The 14B router achieved the strongest results for RAG and Calculator and the highest Balanced Accuracy for Reasoning, while the 7B router obtained the highest Reasoning F1. Web search remained the weakest capability across all router sizes, with low F1 despite comparatively higher balanced accuracy. Overall, the 14B configuration provided the strongest and most consistent performance across the four routing decisions, although the results remained capability-dependent.

\subsection{RQ3: Downstream Quality and Efficiency}
\begin{figure}[H]
    \centering
    \includegraphics[width=0.85\linewidth]{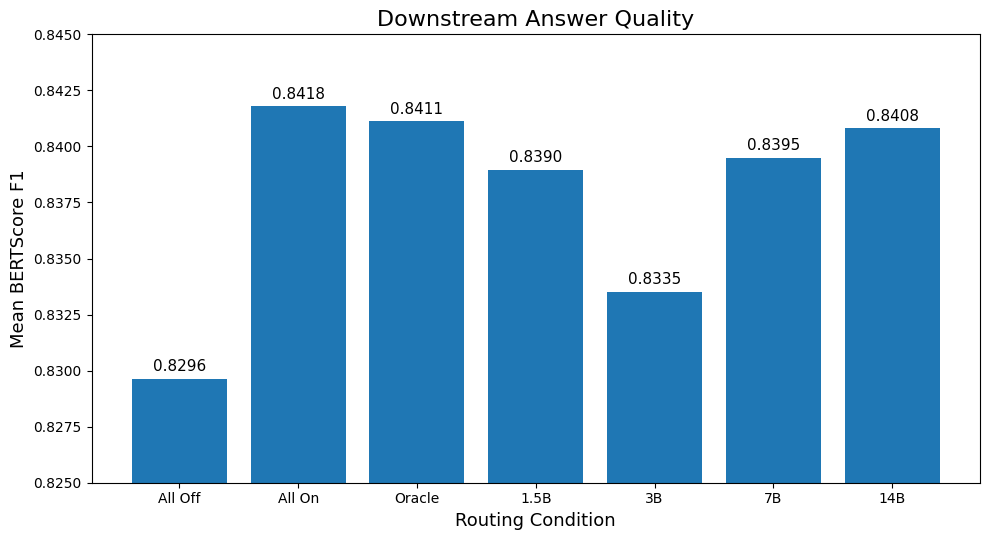}
    \caption{Mean downstream BERTScore F1 across routing conditions.}
    \label{fig:rq3_quality}
    \Description{}
\end{figure}

RQ3 evaluates whether capability routing can reduce downstream computation while maintaining similar reference-answer similarity scores on the benchmark. Figure~\ref{fig:rq3_quality} compares downstream BERTScore F1 across the baseline, oracle, and routed configurations.

\begin{figure}[H]
    \centering
    \includegraphics[width=0.85\linewidth]{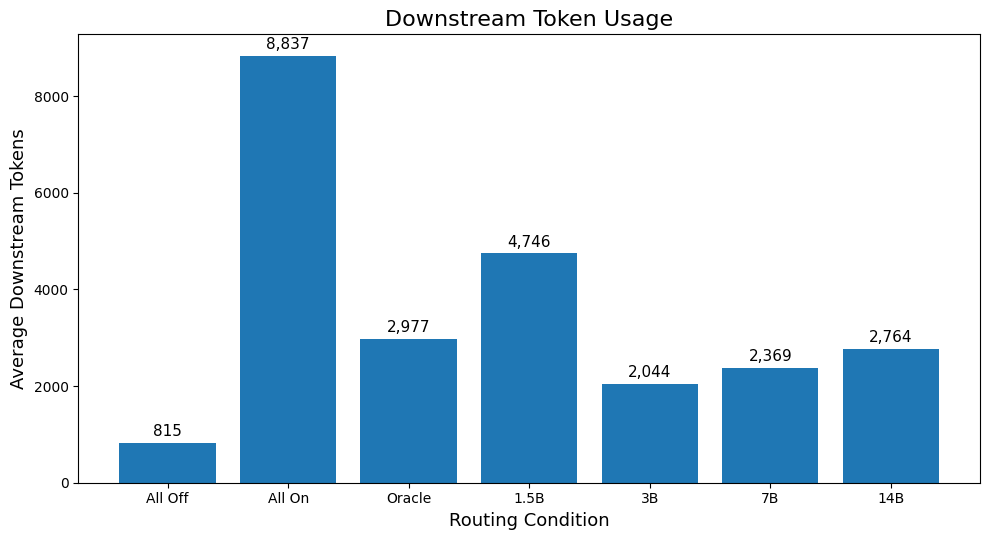}
    \caption{Average downstream token consumption across routing conditions.}
    \label{fig:rq3_tokens}
    \Description{}
\end{figure}

Mean BERTScore F1 values were numerically close across most configurations despite substantial differences in capability activation. ALL\_ON achieved the highest mean BERTScore F1 at 0.8418, followed by ORACLE at 0.8411 and the 14B router at 0.8408. The 7B and 1.5B routers achieved 0.8395 and 0.8390, respectively, while the 3B router produced the lowest routed score at 0.8335. Because BERTScore measures semantic similarity rather than factual, numerical, or task correctness, these values should not be interpreted as demonstrating equivalent or preserved answer quality \cite{bertscore}. The computational differences between these configurations were considerably larger. Figure~\ref{fig:rq3_tokens} shows average downstream token consumption.

\begin{figure}[H]
    \centering
    \includegraphics[width=0.85\linewidth]{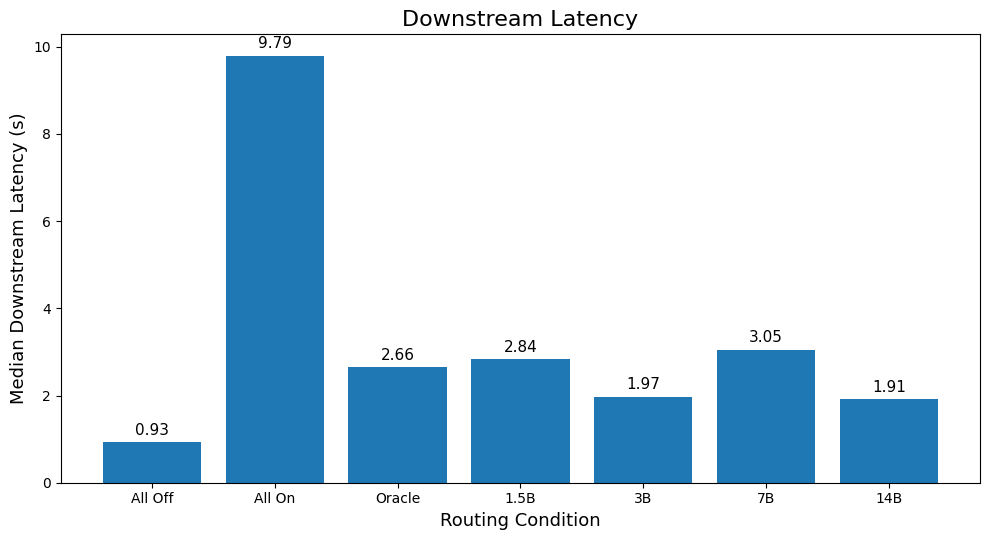}
    \caption{Median downstream latency across routing conditions.}
    \label{fig:rq3_latency}
    \Description{}
\end{figure}

ALL\_ON consumed approximately 8,837 downstream tokens per question, compared with 2,977 for ORACLE and 2,044--4,746 for the routed configurations. Among the routed systems, the 3B router used the fewest downstream tokens at approximately 2,044, followed by 7B at 2,369 and 14B at 2,764. The 1.5B router consumed approximately 4,746 tokens. Thus, numerically similar mean BERTScore values were observed across configurations with substantially different downstream computation. A similar pattern was observed for downstream latency, shown in Figure~\ref{fig:rq3_latency}.
ALL\_ON incurred the highest median downstream latency at 9.79~s, whereas ORACLE required 2.66~s. Among the routed configurations, the 14B router achieved the lowest median downstream latency at 1.91~s, followed by 3B at 1.97~s, 1.5B at 2.84~s, and 7B at 3.05~s. These measurements capture downstream execution only; router inference was performed locally in Google Colab, whereas downstream generation was executed through AWS Bedrock, so the two latency measurements are analyzed separately.

ALL\_OFF also achieved a mean BERTScore F1 that was numerically close to the routed, ORACLE, and ALL\_ON conditions while using substantially fewer downstream tokens. This result suggests that the present evaluation cannot determine whether the optional capabilities provided only limited benefit on this benchmark, whether BERTScore was insufficiently sensitive to the kinds of factual, numerical, or task-level improvements those capabilities were intended to provide, or whether both effects contributed. Accordingly, the observed BERTScore differences should be interpreted descriptively rather than as evidence that the capability configurations were equivalent in answer quality. The differences in the routing error profiles for each model and capability are reported in Table~\ref{tab:rq3_confusion}.

\begin{table}[t]
\centering
\caption{Confusion counts for capability predictions on the
corrected RQ3 benchmark. Unknown and error counts account
for predictions excluded from the confusion matrix.}
\label{tab:rq3_confusion}
\small
\setlength{\tabcolsep}{3pt}
\renewcommand{\arraystretch}{1.1}
\resizebox{\columnwidth}{!}{%
\begin{tabular}{llrrrrrr}
\toprule
\textbf{Router} & \textbf{Capability}
& \textbf{TP} & \textbf{TN}
& \textbf{FP} & \textbf{FN}
& \textbf{Unknown} & \textbf{Errors} \\
\midrule
\multirow{4}{*}{1.5B}
& Reasoning  & 44 & 25 & 18 & 39 & 0 & 1 \\
& RAG        & 40 & 27 & 3  & 55 & 2 & 0 \\
& Web Search & 9  & 62 & 55 & 1  & 0 & 0 \\
& Calculator & 27 & 31 & 65 & 2  & 2 & 0 \\
\midrule
\multirow{4}{*}{3B}
& Reasoning  & 46 & 37  & 7 & 37 & 0 & 0 \\
& RAG        & 19 & 30  & 0 & 78 & 0 & 0 \\
& Web Search & 1  & 114 & 3 & 9  & 0 & 0 \\
& Calculator & 16 & 93  & 3 & 13 & 2 & 0 \\
\midrule
\multirow{4}{*}{7B}
& Reasoning  & 57 & 30  & 14 & 26 & 0 & 0 \\
& RAG        & 54 & 30  & 0  & 43 & 0 & 0 \\
& Web Search & 1  & 111 & 6  & 9  & 0 & 0 \\
& Calculator & 19 & 98  & 0  & 10 & 0 & 0 \\
\midrule
\multirow{4}{*}{14B}
& Reasoning  & 48 & 41 & 3  & 35 & 0 & 0 \\
& RAG        & 92 & 26 & 4  & 5  & 0 & 0 \\
& Web Search & 3  & 99 & 18 & 7  & 0 & 0 \\
& Calculator & 22 & 97 & 1  & 7  & 0 & 0 \\
\bottomrule
\end{tabular}%
}
\end{table}

The router-level results help explain the downstream behavior observed in Figures~\ref{fig:rq3_quality}--\ref{fig:rq3_latency}. Although the 14B router was generally strongest across capabilities, smaller routers were competitive on individual tasks. The 7B router achieved the highest Reasoning F1, while the 1.5B router achieved the highest Web Search F1. However, similar classification scores did not necessarily produce similar execution costs. The confusion counts show that the 1.5B router produced 55 Web Search false positives and 65 Calculator false positives, causing frequent unnecessary capability activation. Although its mean BERTScore F1 was numerically close to that of the 14B router (0.8390 vs.\ 0.8408), it required substantially more downstream tokens (4,746 vs.\ 2,764) and higher median downstream latency (2.84~s vs.\ 1.91~s). The 7B router similarly obtained a mean BERTScore F1 close to 14B (0.8395 vs.\ 0.8408), but with higher median downstream latency (3.05~s) and slightly lower token consumption (2,369 tokens). The 3B router was more conservative, producing relatively few false positives and the lowest downstream token consumption among routed configurations (2,044 tokens), but it also produced many false negatives, including 78 missed RAG activations, and the lowest routed mean BERTScore F1 (0.8335).

Among the routed configurations, the 14B router achieved the highest mean BERTScore F1 and the lowest median downstream latency, while consuming fewer downstream tokens than the 1.5B router but more than the 3B and 7B routers. These results describe performance on the reported downstream metrics only and do not establish an overall optimum in total compute, monetary cost, or end-to-end latency. Its relatively strong RAG and calculator-routing performance coincided with the highest mean BERTScore F1 among learned routers and the lowest median downstream latency. These findings indicate that router performance cannot be interpreted from F1 or balanced accuracy alone: the distribution of false positives and false negatives directly affects which capabilities execute and therefore influences downstream token consumption and latency. The observed BERTScore differences remain descriptive and do not establish statistical equivalence or correctness across routing conditions.
\begin{figure}[H]
    \centering
    \includegraphics[width=0.75\linewidth]{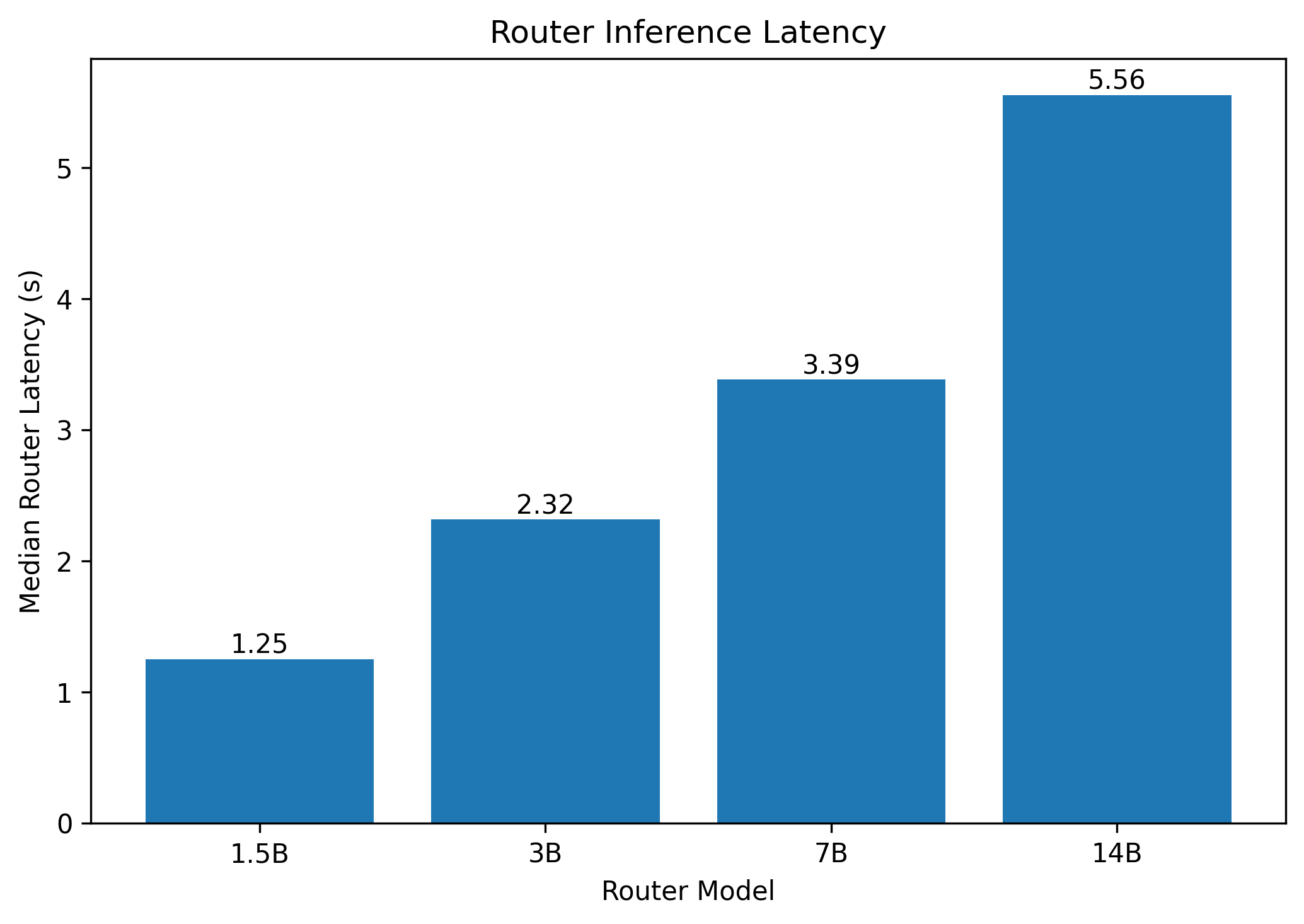}
    \caption{Median inference latency of the four SLM router configurations.}
    \label{fig:rq3_router_latency}
    \Description{}
\end{figure}
Router inference latency increased with model capacity, as shown in Figure~\ref{fig:rq3_router_latency}. Median router latency increased from approximately 1.25~s for the 1.5B router to 5.56~s for the 14B router. Thus, although the 14B router reduced downstream execution latency through more selective capability activation, the router itself was more computationally expensive. Router and downstream latency are reported separately because they were measured under different inference environments. The Qwen2.5 routers were executed through Ollama in Google Colab, whereas the 120B downstream model was accessed through AWS Bedrock. Directly summing these measurements would therefore not provide a controlled end-to-end latency comparison. Future work should evaluate both components in a common inference environment to measure total system latency more rigorously.

\section{Discussion}
\label{sec:discussion}

Router capacity and prompt design jointly influenced
capability classification performance. Under minimal
prompting, increasing model size did not consistently
improve classification performance. With capability-specific
prompt optimization, however, the 14B router achieved the
highest F1 across all four tasks. The extent of this
advantage varied across capabilities: smaller routers
remained competitive for reasoning and calculator routing,
whereas larger performance gaps persisted for RAG and
web search. Prompt optimization therefore compensated
for reduced router capacity more effectively for some
capability decisions than for others.

The most effective prompting strategy also depended on
the capability. GEPA produced the strongest results for
RAG and web search, while DSPy Few-Shot was generally
strongest for reasoning. Calculator classification
remained competitive under simpler Minimal Baseline or
Descriptive SME prompts. More elaborate prompting was
therefore not consistently associated with improved
performance. Furthermore, optimized prompts sometimes
increased invalid outputs, particularly for the 1.5B
router. Classification scores must therefore be
interpreted alongside output reliability, as a
configuration that performs well on valid predictions
may still require frequent fallback handling during
deployment.

The downstream evaluation demonstrates why classification
performance alone is insufficient for router selection.
The 1.5B and 14B configurations achieved numerically
similar mean BERTScore F1 values, but their downstream
execution requirements differed substantially. The 1.5B
router produced numerous false-positive web search and
calculator decisions and consumed approximately 4,746
downstream tokens per question, compared with 2,764
for the 14B router. Its median downstream latency was
also higher (2.84 versus 1.91 seconds). These results
associate unnecessary capability activation with increased
downstream computation, although a positive calculator
decision only made the tool available and did not
necessarily result in an invocation.

The 3B configuration illustrated a different trade-off.
It consumed the fewest downstream tokens, approximately
2,044 per question, but also achieved the lowest mean
BERTScore F1 among the routed configurations. Its
conservative predictions included numerous missed RAG
activations. Lower token consumption therefore did not,
by itself, indicate more effective routing. Because
individual questions could contain errors across
multiple capabilities, these associations do not isolate
the causal effect of any individual routing error.

Among the routed configurations, the 14B router achieved
the highest mean BERTScore F1 and the lowest median
downstream latency. Its mean BERTScore F1 was 0.8408,
compared with 0.8418 for ALL\_ON and 0.8411 for ORACLE.
Relative to ALL\_ON, average downstream token consumption
decreased from approximately 8,837 to 2,764, representing
a reduction of 68.7\%, while median downstream latency
decreased from 9.79 to 1.91 seconds, representing a
reduction of 80.5\%. Thus, on the evaluated benchmark,
substantial reductions in downstream computation were
accompanied by only a small numerical difference in
reference-answer similarity.

\begin{sloppypar}
These findings require two qualifications. First,
BERTScore measures semantic similarity to reference
answers and does not establish factual, numerical, or
regulatory correctness \cite{bertscore}. The small differences
in mean scores also do not demonstrate statistical
equivalence. Second, the reported downstream savings exclude
router overhead, which increased with router capacity.
Router and downstream latency were measured separately in
Google Colab and AWS Bedrock, respectively. The results
therefore characterize downstream computational savings
and routing overhead under the evaluated conditions,
rather than establishing an overall latency advantage
for the 14B configuration.
\end{sloppypar}

The scope of the downstream benchmark further limits
generalization. The benchmark contains 127 questions
concerning pharmaceutical regulation, of which only
ten are labeled as requiring web search. Broader workloads
and task-specific correctness assessments are needed
to determine whether the observed trade-offs persist
beyond this setting. Evaluating the complete pipeline
under a controlled deployment would also clarify when
downstream savings outweigh the additional computational
cost of a larger router.

Taken together, the results support capability-specific
router selection. Router capacity, prompting strategy,
output reliability, and error profiles each influence
the resources ultimately consumed by the downstream
system. A smaller router may be advantageous when its
reduced overhead is accompanied by sufficiently effective
capability decisions; classification F1 alone cannot
establish this balance.

\section{Conclusion}
\label{sec:conclusion}

This study evaluated prompt-driven small language models
as pre-inference routers for additional reasoning effort,
RAG, web search, and calculator access. The experiments
combined a within-family comparison of router capacity
and prompting strategies with an evaluation of downstream
execution using a fixed language model.

For RQ1, increasing router capacity under minimal prompting
did not consistently improve classification performance.
For RQ2, prompt optimization altered this relationship,
with the 14B router achieving the highest optimized F1
across all four capabilities. Smaller routers approached
its performance more closely for reasoning and calculator
routing than for RAG and web search, indicating that the
extent of prompt-based compensation depends on the
capability.

For RQ3, selective routing substantially reduced downstream
token consumption and latency while producing mean
BERTScore F1 values numerically close to those of ALL\_ON
and ORACLE. The 14B configuration achieved the highest
mean BERTScore F1 and the lowest median downstream latency
among the evaluated routers. However, its greater routing
overhead prevents these results from establishing it as
the best overall system configuration. The contrasting
behavior of the 1.5B and 3B routers further demonstrated
that similar classification scores or lower token consumption
do not necessarily imply equivalent downstream performance.

The contribution of this study is an empirical evaluation
linking router capacity and prompt design to capability
selection and downstream computation. The findings
demonstrate where smaller, prompt-optimized routers remain
competitive and why their suitability must also be
assessed through output reliability, error profiles,
and routing overhead. 

Although the present study provides valuable insights into the effectiveness of the proposed approach, there are several opportunities for further investigation. Future work will focus on extending the evaluation to more diverse datasets encompassing different domains, tasks, and data distributions to assess the generalizability and robustness of the findings. Furthermore, the study will be expanded to include a broader range of language models with fewer than 32 billion parameters, covering different model architectures, parameter scales, and training strategies. Such an investigation will help determine whether the observed trends are consistent across different model families and sizes. Additionally, exploring more comprehensive evaluation settings and benchmarking strategies will provide a deeper understanding of the applicability, limitations, and performance characteristics of the proposed approach.

\bibliographystyle{ACM-Reference-Format}
\bibliography{references}

\appendix
\section{Detailed Experimental Configuration}
\label{app:experimental_configuration}

This appendix documents the principal software, model, prompting, retrieval, tool-use, and evaluation settings used in the experiments. Settings not explicitly recorded during execution are reported as unavailable rather than reconstructed retrospectively.

\begin{table*}[t]
\centering
\small
\setlength{\tabcolsep}{5pt}
\renewcommand{\arraystretch}{1.12}
\caption{Experimental software, model, and inference configuration.}
\label{tab:experimental_configuration}
\begin{tabularx}{\textwidth}{@{}>{\raggedright\arraybackslash}p{0.23\textwidth} >{\raggedright\arraybackslash}X@{}}
\toprule
\textbf{Component} & \textbf{Configuration} \\
\midrule
Router models &
\nolinkurl{qwen2.5:1.5b}, \nolinkurl{qwen2.5:3b}, \nolinkurl{qwen2.5:7b}, and \nolinkurl{qwen2.5:14b} \cite{qwen25}, served through Ollama. \\

Ollama &
Version 0.40.4; local API endpoint \nolinkurl{127.0.0.1:11434}. \\

DSPy &
Version 3.4.1 \cite{dspy}.  \\

Raw router inference &
Temperature = 0.0; maximum generated tokens (\nolinkurl{num_predict}) = 10. \\

DSPy router inference &
Temperature = 0.0; maximum generated tokens = 64. \\

Few-Shot construction &
Thirty examples were drawn from the 100-example training set to form the Few-Shot candidate pool. The DSPy configuration permitted up to 6 labeled demonstrations, up to 4 bootstrapped demonstrations, and 4 candidate programs. \\

GEPA optimizer &
\nolinkurl{dspy.GEPA} \cite{gepa} initialized from the DSPy Few-Shot program, with \nolinkurl{auto="light"}, \nolinkurl{num_threads=1}, \nolinkurl{track_stats=True}, and random seed 42. \\

GEPA reflection model &
GPT-OSS-120B, using \nolinkurl{openai.gpt-oss-120b-1:0}. \\

RAG embedding model &
\nolinkurl{sentence-transformers/all-MiniLM-L6-v2} \cite{sentencebert}. \\

RAG chunking &
Chunk size = 800; chunk overlap = 150. \\

RAG retrieval &
FAISS vector retrieval \cite{faiss} with top-$k=4$.\\
\bottomrule
\end{tabularx}
\end{table*}

\begin{table*}[t]
\centering
\small
\setlength{\tabcolsep}{5pt}
\renewcommand{\arraystretch}{1.12}
\caption{Experimental software, model, and inference configuration (continued).}
\label{tab:experimental_configuration_continued}
\begin{tabularx}{\textwidth}{@{}>{\raggedright\arraybackslash}p{0.23\textwidth} >{\raggedright\arraybackslash}X@{}}
\toprule
\textbf{Component} & \textbf{Configuration} \\
\midrule
Web Search &
Tavily search with \nolinkurl{search_depth="advanced"}, maximum 5 results, and \nolinkurl{include_answer=False}. \\

Calculator &
Structured calculator tool exposed to the downstream model through a JSON tool schema requiring a single arithmetic \nolinkurl{expression} string. Expressions were evaluated with a restricted Python AST evaluator. \\

Downstream model &
\nolinkurl{openai.gpt-oss-120b-1:0} through AWS Bedrock. \\

Downstream generation &
Temperature = 0.0; maximum generated tokens = 2048 for the standard downstream configuration. \\

BERTScore &
BERTScore \cite{bertscore} with \nolinkurl{roberta-large}; final scoring was executed on GPU. Additional options not explicitly retained in the experiment metadata are reported as unavailable. \\

Router hardware &
Google Colab with an NVIDIA T4 GPU. Other Colab hardware characteristics were not fixed and are therefore not reported. \\

Downstream hardware &
Managed AWS Bedrock infrastructure; underlying server hardware was not exposed to the experiment and is unavailable. \\

Routing-call execution & The four capability-routing calls were executed sequentially for each question. Router predictions were generated and stored first, and the downstream evaluation was performed subsequently using the saved routing decisions. \\
\bottomrule
\end{tabularx}
\end{table*}

\subsection{Representative Prompt Configurations}

The complete model- and method-specific router prompts and selected Few-Shot demonstrations were retained with the experimental artifacts. To avoid reproducing several pages of near-duplicate prompts, the downstream prompt and representative optimized router instructions are presented here.

\subsubsection{RQ3 Downstream Prompt}

The downstream model used the following fixed system instruction across all RQ3 conditions:

\begin{routerprompt}
\small
\texttt{You are a precise question-answering assistant.}

\texttt{Answer the user's question using the provided information and tools.}

\texttt{If retrieved context is provided, use it where relevant.}
\texttt{If web-search context is provided, use it where relevant.}
\texttt{If a calculator result is provided, use that result directly.}

\texttt{Give the final answer clearly and directly.}
\end{routerprompt}

The system prompt was supplemented by a fixed tool policy specifying that RAG and web search were performed externally, that the calculator was the only callable tool, and that it should be used when arithmetic was required. The user message contained the original question, followed by retrieved-document and web-search context when the corresponding capabilities were enabled.

\subsection{Representative Router Prompts}

Five prompting strategies were evaluated: Minimal Baseline, Descriptive SME, DSPy Baseline, DSPy Few-Shot, and GEPA. To keep the appendix concise, one representative prompt from each strategy is reproduced below. Complete model--capability prompt configurations and selected Few-Shot demonstrations were retained with the experimental artifacts.

\subsubsection{Minimal Baseline Example: Reasoning}

\begin{routerprompt}
\small
Classify the Input into exactly one category: ``REASONING REQUIRED'' or ``NO REASONING''.

CONSTRAINT: Output ONLY one of these exact strings:

\texttt{REASONING REQUIRED}

\texttt{NO REASONING}
\end{routerprompt}

\subsubsection{Descriptive SME Example: Calculator}

\begin{routerprompt}
\small
You are a binary classifier for pharmaceutical prompts.
Classify the Input into exactly one category: ``CALCULATOR REQUIRED'' or ``NO CALCULATOR''.

\textbf{LOGIC RULES:}

1. Label as ``CALCULATOR REQUIRED'' IF the prompt asks to:
\begin{itemize}
    \item Perform arithmetic or numerical computation.
    \item Calculate percentages, ratios, averages, rates, doses, concentrations, or conversions.
    \item Apply a mathematical formula to numerical values.
    \item Derive a numerical result from values provided in the prompt.
\end{itemize}

2. Label as ``NO CALCULATOR'' IF the prompt asks to:
\begin{itemize}
    \item Define or explain a term or concept.
    \item Retrieve or state factual information.
    \item Perform qualitative reasoning without numerical computation.
    \item Extract or identify numerical information without calculating anything.
\end{itemize}

CONSTRAINT: Output ONLY the category name. No other text.
\end{routerprompt}

\subsubsection{DSPy Baseline Example: RAG}

\begin{routerprompt}
\small
Classify the Input into exactly one category: ``RAG REQUIRED'' or ``NO RAG''.

CONSTRAINT: Output ONLY one of these exact strings:

\texttt{RAG REQUIRED}

\texttt{NO RAG}
\end{routerprompt}

\subsubsection{DSPy Few-Shot Example: Calculator}

The following DSPy Few-Shot configuration was used for the Qwen2.5-1.5B Calculator router:

\begin{routerprompt}
\small
Classify the Input into exactly one category: ``CALCULATOR REQUIRED'' or ``NO CALCULATOR''.

CONSTRAINT: Output ONLY one of these exact strings:

\texttt{CALCULATOR REQUIRED}

\texttt{NO CALCULATOR}

\vspace{0.5em}
\textbf{Selected demonstrations:}

\textbf{Input:} How would excipient functionality influence scale-up decisions?

\textbf{Label:} \texttt{NO CALCULATOR}

\vspace{0.3em}

\textbf{Input:} Ten dosage units have a mean content of 96.33\% of label claim and a sample standard deviation of 3.31 percentage points. Calculate the coefficient of variation as a percentage.

\textbf{Label:} \texttt{CALCULATOR REQUIRED}

\vspace{0.3em}

\textbf{Input:} List two common dosage forms developed during formulation development.

\textbf{Label:} \texttt{NO CALCULATOR}

\vspace{0.3em}

\textbf{Input:} Calculate the effective annual rate (EAR) for a nominal rate of 6\% compounded quarterly.

\textbf{Label:} \texttt{CALCULATOR REQUIRED}

\vspace{0.3em}

\textbf{Input:} Analyze this network packet capture file to identify potential port scanning activities.

\textbf{Label:} \texttt{CALCULATOR REQUIRED}

\vspace{0.3em}

\textbf{Input:} Two analytical methods report results of 90.6 and 112.06 units for the same sample. Using the first result as the reference, calculate the percentage difference of the second result.

\textbf{Label:} \texttt{CALCULATOR REQUIRED}
\end{routerprompt}

\subsubsection{GEPA-Optimized Example: Web Search}

The following GEPA-optimized prompt was used for the Qwen2.5-3B Web Search router:

\begin{routerprompt}
\small
Given an \nolinkurl{input_text}, decide whether answering it requires up-to-date information that can change over time or depends on current conditions. Output \textbf{exactly} one of the two strings -- \texttt{WEB SEARCH REQUIRED} or \texttt{NO WEB SEARCH} -- with no extra characters, whitespace, or explanation.

\textbf{When to output \texttt{WEB SEARCH REQUIRED}}
\begin{itemize}
    \item The query asks for \emph{current} or \emph{recent} data that is likely to vary: availability, stock levels, pricing, market share, sales figures, quarterly or annual statistics, recent rankings, current policies, regulations, travel advisories, visa requirements, news events, clinical trial results, weather forecasts, COVID-19 case counts, etc.
    \item The question contains temporal cues such as ``current'', ``latest'', ``today'', ``now'', ``this week/month/quarter/year'', ``Q1/Q2/...'', ``2023/2024'', ``as of'', ``recently'', ``latest update'', ``2024-05-01'', or any request for ``percentage'', ``share'', ``ranking'', ``trend'', ``forecast'', ``price'', ``availability'', ``stock'', ``inventory'', ``number of ...'', ``how many ... now'', etc.
    \item The answer cannot be reliably derived from static encyclopedic or scientific knowledge because it depends on real-time sources.
\end{itemize}

\textbf{When to output \texttt{NO WEB SEARCH}}
\begin{itemize}
    \item The query can be answered using stable, well-known facts that do not change over time: chemical composition, standard pharmacological definitions, established scientific principles, historical events, classic literature, mathematical formulas, long-standing legal definitions, generic medical information, basic geography, etc.
    \item No temporal or location-specific cue that would require the latest data is present.
\end{itemize}

\textbf{Output format}
\begin{itemize}
    \item Print only the chosen label (\texttt{WEB SEARCH REQUIRED} or \texttt{NO WEB SEARCH}).
    \item Do not add any punctuation, spaces, newline characters, or additional text before or after the label.
\end{itemize}

Follow these rules for every \nolinkurl{input_text}.
\end{routerprompt}
\end{document}